\documentclass{article}

     \PassOptionsToPackage{numbers, compress}{natbib}

     \usepackage[preprint]{neurips_2023}

\usepackage[utf8]{inputenc} 
\usepackage[T1]{fontenc}    
\usepackage{hyperref}       
\usepackage{url}            
\usepackage{booktabs}       
\usepackage{amsfonts}       
\usepackage{nicefrac}       
\usepackage{microtype}      
\usepackage{xcolor}         
\usepackage{amsmath}        
\usepackage{graphicx}     
\usepackage{appendix}       
\usepackage{enumerate}      
\usepackage{multirow}        
\usepackage{threeparttable}   

\title{Cross-LKTCN:\\Modern Convolution Utilizing Cross-Variable Dependency for Multivariate Time Series Forecasting}

\author{%
	Donghao Luo, Xue Wang\\
	Department of Precision Instrument, Tsinghua University, China\\
	\texttt{ldh21@mails.tsinghua.edu.cn}\\
}

\begin{document}

\maketitle

\begin{abstract}
 The past few years have witnessed the rapid development in multivariate time series forecasting. The key to accurate forecasting results is capturing the long-term dependency between each time step (cross-time dependency) and modeling the complex dependency between each variable (cross-variable dependency) in multivariate time series. However, recent methods mainly focus on the cross-time dependency but seldom consider the cross-variable dependency. To fill this gap, we find that convolution, a traditional technique but recently losing steam in time series forecasting, meets the needs of respectively capturing the cross-time and cross-variable dependency. Based on this finding, we propose a modern pure convolution structure, namely Cross-LKTCN, to better utilize both cross-time and cross-variable dependency for time series forecasting. Specifically in each Cross-LKTCN block, a depth-wise large kernel convolution with large receptive field is proposed to capture cross-time dependency, and then two successive point-wise group convolution feed forward networks are proposed to capture cross-variable dependency. Experimental results on real-world benchmarks show that Cross-LKTCN achieves state-of-the-art forecasting performance and improves the forecasting accuracy significantly compared with existing convolutional-based models and cross-variable methods.
\end{abstract}

\section{Introduction}
\label{intro}

Multivariate time series records $M$ variables’ variations during multiple time steps. It mainly contains two types of dependencies, namely cross-time dependency and cross-variable dependency. The former indicates the long-term temporal dependency between each time step. The latter is the dependency among different variables. And time series forecasting means using the historical information of length $L$  to predict the future of length $T$, where $T$ represents a long-term period of time. It has been widely used in industrial production planning, sensor network monitoring and health management. 

The past few years have witnessed the rapid development in time series forecasting\cite{wen2022transformers,lim2021time}. Among them, the rise of Transformer-based methods \cite{nie2022time,wu2021autoformer,zhou2021informer,zhou2022fedformer,zhang2023crossformer,li2019enhancing,cirstea2022triformer,vaswani2017attention,kitaev2020reformer,liu2021pyraformer} and Linear models \cite{zeng2022transformers} is especially compelling. However, \textbf{the cross-variable dependency has been seldom considered in these previous works}. Most previous models simply mix and embed $M$ variables at a specific time step into a $D$-dimensional vector, trying to capture the cross-variable dependency via the embedding layer \cite{wu2021autoformer,zhou2021informer,zhou2022fedformer,wu2022timesnet}. However, just an embedding projection layer fails to learn the complex dependency across variables and even loses their independent characteristics for not considering the different behaviors of different variables. Other models embed and analyse each univariate time series individually to maintain the independence of different variables but still omit the dependency between them \cite{nie2022time,zeng2022transformers,cirstea2022triformer,zhou2022film}. Some cross-variable methods design explicit mechanisms to capture cross-variable dependency \cite{wu2020connecting,zhang2023crossformer,lai2018modeling}. But they come with high computational complexity. More efficient methods to capture both cross-variable and cross-time dependency are still in need.

From the perspective of computational complexity, convolution\footnote{The convolution in this paper refers to 1D convolution by default, unless specified as 2D convolution.} is an efficient way to capture cross-variable dependency\cite{lai2018modeling}. Meanwhile, the decoupling property of group convolution and depth-wise separable convolution\cite{howard2017mobilenets,krizhevsky2017imagenet,zhang2018shufflenet} meets the needs of respectively capturing cross-time dependency and cross-variable dependency. However, convolution-based models are losing steam in time series forecasting due to their limited receptive field and weak cross-time dependency modeling ability \cite{bai2018empiricaltcn1}. Although some methods \cite{wangmicn,liu2022scinet} use multi-resolution down-sampling to alleviate this problem, there is still a performance gap between them and the state-of-the-art models with global view \cite{zeng2022transformers,nie2022time}. A turnaround occurred recently. In computer vision, it is proven that large kernel 2D convolution-based models can obtain as large receptive field as vision transformers have\cite{liu2022convnet,ding2022scaling}. Some recipes for training the large kernel convolutions are provided in \cite{ding2022scaling}. Therefore we can also enlarge the kernel size of 1D convolution-based model in time series forecasting to get rid of the limited receptive field.

Based on above motivations, we bring convolution-based model back to the arena of time series forecasting and propose Cross-LKTCN as a modern pure convolution structure\footnote{A modern pure convolution structure means a convolution-based model incorporating some architectural designs in Transformers but free of any attention mechanism\cite{liu2022convnet}.} to efficiently utilize cross-time and cross-variable dependency for time series forecasting. Specifically, we introduce the patch-style embedding strategy\footnote{Patch-style embedding strategy means doing patching and embedding on the input series, while the commonly used non-patch-style embedding strategy only embeds the input series without patching process.} for time series to enhance locality and aggregate more semantic information from adjacent time points. Then we propose the Cross-LKTCN block, which contains a depth-wise large kernel convolution to capture the cross-time dependency and two successive point-wise group convolution feed forward networks (FFNs) to capture cross-variable dependency. And we remove causal convolution from our design as it was proven unnecessary in time series forecasting \cite{liu2022scinet}. Experimentally, Cross-LKTCN achieves competitive performance on nine real-world benchmarks against state-of-the-arts, indicating the great potential of convolution-based models and cross-variable methods in time series forecasting. \textbf{Our contributions are as follows:}

\begin{itemize}
	\item To better utilize the cross-variable dependency in multivariate time series forecasting, we propose successive point-wise group convolution FFNs based on the decoupling property of group convolution and depth-wise separable convolution. Our method surpasses existing cross-variable methods and achieves state-of-the-art performance, indicating the importance of cross-variable dependency in time series forecasting.
	\item Inspired by the latest large kernel 2D convolution in computer vision, we introduce the depth-wise large kernel convolution into time series forecasting to enlarge the receptive field and improve the cross-time dependency modeling ability. Experimental results show that our method can better unleash the potential of convolution in time series forecasting than other existing convolution-based models.
	\item Extensive experimental results on nine real-world benchmarks show the effectiveness of our Cross-LKTCN. Our method surpasses existing convolution-based models and cross-variable methods by a large margin in multivariate time series forecasting (respectively 27.4\% and 52.3\% relative improvement) and achieves state-of-the-art performance.
\end{itemize}

\section{Related Work}
\label{Related_work}

\subsection{Utilizing Cross-variable Dependency in Multivariate Time Series}
Unlike the \textbf{variable-mixing methods} \cite{wu2021autoformer,zhou2021informer,zhou2022fedformer,wu2022timesnet} and \textbf{variable-independent methods} \cite{nie2022time,zeng2022transformers,cirstea2022triformer,zhou2022film} mentioned in Section \ref{intro}, some cross-variable methods \textbf{focus on capturing cross-variable dependency} and design some explicit mechanisms. LSTnet\cite{lai2018modeling} employs convolution to capture cross-variable dependency and uses the recurrent component\cite{hochreiter1997long,chung2014empirical} to capture cross-time dependency. This study proves the effectiveness of convolution in capturing cross-variable dependency. However, following studies like MTGNN\cite{wu2020connecting} and Crossformer\cite{zhang2023crossformer} omit to further optimize the efficiency and performance of convolution but turn to design dedicated graph convolution or attention mechanism to capture the cross-variable dependency. These cross-variable methods can explicitly capture cross-variable dependency. But they come with high computational complexity and are still not comparable to the state-of-the-art variable-independent methods\cite{zeng2022transformers,nie2022time} in long-term forecasting. More efficient designs for capturing cross-variable dependency are still in need.

\subsection{Capturing Cross-time Dependency via Convolution in Time Series Forecasting}
Convolution is widely used in time series forecasting. TCN\cite{bai2018empiricaltcn1} uses causal convolution to model the temporal causality and stacks many layers for larger receptive field. Similar ideas are also adopted by  \cite{borovykh2017conditional,sen2019think,gu2021combining}.   Going beyond causal convolution, MICN\cite{wangmicn} proposes a multi-scale branch structure with down-sampled convolution and isometric convolution to combine local features and global correlations in time series. However, these pure convolution-based models haven't fully unleashed convolution's potential in time series for not adopting modern convolution structures. Considering the limited receptive field, many models only introduce convolution to enhance locality. LogTrans\cite{li2019enhancing} introduces convolution to Transformer to enhance locality and uses LogSparse attention for long-term dependency modeling. SCINet\cite{liu2022scinet} removes the idea of causal convolution and introduces a recursive downsample-convolve-interact architecture to model time series with complex temporal dynamics. But they still have difficulty in modeling long-term dependency due to the limited receptive filed.

\subsection{Large Kernel 2D Convolution in Computer Vision}
Large kernel 2D convolution has a long history in computer vision but was abandoned for a long time. It is brought back in some modern ConvNets nowadays and proved to be an effective way to enlarge the receptive field.
AlexNet\cite{krizhevsky2017imagenet} uses a large kernel size such as 7×7 and 11×11 in 2010s. However, with the introduction of VGG\cite{simonyan2014very}, it has become a consensus to stack small convolutional kernels such as 1×1 and 3×3 to obtain large receptive field\cite{he2015deep,huang2017densely}. In 2020s, Vision Transformers (ViTs)\cite{dosovitskiy2020image,liu2021swin} are proposed and outperform previous standard ConvNets\cite{he2015deep,xie2017aggregatedresnext}. Inspried by the architectural designs in ViTs, modern ConvNets in 2020s are introduced. ConvNeXt\cite{liu2022convnet} adopts a depth-wise separable 7×7 kernel, surpassing the performance of Swin Transformer\cite{liu2021swin}. Further more, RepLKNet\cite{ding2022scaling} scales the kernel size to 31×31 with the help of Structural Reparameter technique. SLaK\cite{liu2022more} enlarges the kernel size to 51×51 by decomposing a large kernel into two rectangular, parallel kernels and by using dynamic sparsity. Inspired by above studies, we also apply large kernel 1D convolution in time series forecasting to enlarge the receptive field and improve cross-time dependency modeling abiltiy.

\section{Cross-LKTCN}
\label{section3}
\begin{figure}[htb]
	\centering 
	\includegraphics[width=0.95\textwidth]{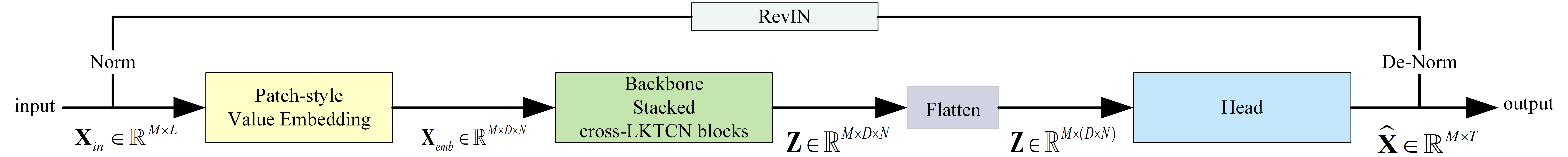}
	\caption{Cross-LKTCN. The patch-style embedding method is applied to time series to enhance locality. The backbone stacked by Cross-LKTCN blocks is proposed to learn the representation by capturing both cross-time and cross-variable dependency. The linear head with a flatten layer is used to obtain the final predictions. And RevIN \cite{kim2021reversible} is to mitigate the distribution shift.}	   
	\label{Fig.1} 
\end{figure}

Time series forecasting aims to predict the future values of prediction length $T$ based on the historical values of input length $L$. We design Cross-LKTCN as a pure convolution structure to utilize both cross-time and cross-variable dependency for time series forecasting. In Section \ref{sec31}, we introduce the overall structure of Cross-LKTCN. In Section \ref{sec32}, we introduce the patch-style embedding strategy for time series to enhance locality and aggregate more local semantic information by patching the adjacent time points. In Section \ref{sec33}, we propose the Cross-LKTCN block, which can capture long-term temporal dependency by depth-wise large kernel convolution and capture cross-variable dependency based on successive point-wise group convolution FFNs.

\subsection{Overall Structure}
\label{sec31}
\paragraph{Forward Process.}The overall structure of Cross-LKTCN is shown on Figure \ref{Fig.1}. We denote ${\mathbf{X}_{in}}\in {{\mathbb{R}}^{M\times L}}$ as the $M$ variables input time series of length $L$. And it will be further divided into $N$ patches and embedded into $D$-dimensional embedding vectors by using patch-style embedding strategy. The embedding process is shown as follows:
\begin{equation}\label{eq1}
\mathbf{X} _{emb}   =\mathrm{Embedding}(\mathbf{X} _{in})   
\end{equation}
Where $\mathrm{Embedding}(\cdot )$ is the patch-style embedding method designed for time series. Details will be introduced in Section \ref{sec32}. After the embedding process, we have the input embedding ${\mathbf{X}_{emb}}\in {{\mathbb{R}}^{M\times D\times N}}$.

Then the input embedding vector ${\mathbf{X}_{emb}}$ is fed into the backbone to capture both the cross-time and cross-variable dependency and to learn the informative representation $\mathbf{Z}\in {{\mathbb{R}}^{M\times D\times N}}$ :
\begin{equation}\label{eq2}
\mathbf{Z}=\mathrm{Backbone}({\mathbf{X}_{emb}})
\end{equation}
$\mathrm{Backbone}(\cdot )$ is the stacked Cross-LKTCN blocks we proposed and will be described in Section \ref{sec33}.

Finally, the linear head with a flatten layer is used to obtain the final prediction:
\begin{equation}\label{eq3}
\widehat{\mathbf{X}}=\mathrm{Head}(\mathrm{Flatten}(\mathbf{Z}))
\end{equation}

Where $\widehat{\mathbf{X}}\in {{\mathbb{R}}^{M\times T}}$ is the prediction of length $T$ with $M$ variables. $\mathrm{Flatten}(\cdot )$ denotes a flatten layer that changes the final representation's shape to $\mathbf{Z}\in {{\mathbb{R}}^{M\times (D\times N)}}$.  $\mathrm{Head}(\cdot )$ indicates the linear projection layer that maps the final representation to the final prediction.

\paragraph{RevIN.}
RevIN\cite{kim2021reversible} is a special instance normalization for time series to mitigate the distribution shift between the training and testing data. In norm phase, we normalize the input time series per variable with zero mean and unit standard deviation before patching and embedding. Then in de-norm phase, we add the mean and deviation back to the final prediction per variable after the forward process.

\subsection{Patch-style Embedding Strategy}
\label{sec32}
\begin{figure}[htb]
	\centering 
	\includegraphics[width=0.95\textwidth]{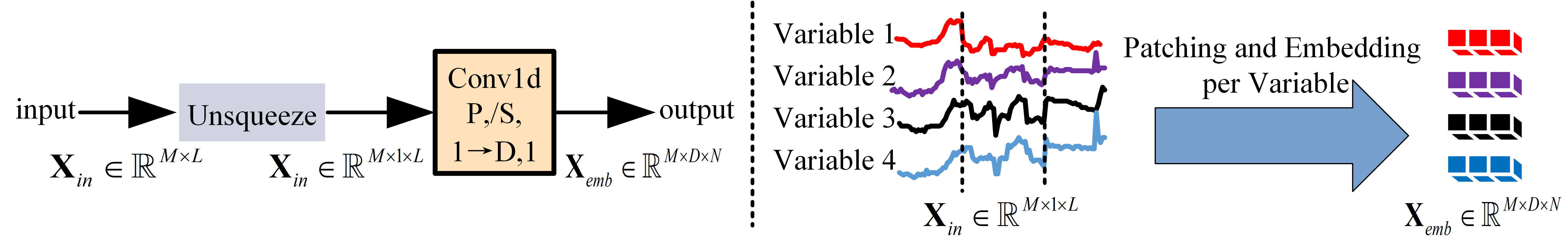}
	\caption{Patch-style Embedding. It employs a 1D convolution layer with kernel size $P$ and stride $S$ to divide the length-$L$ time series into $N$ patches and embed them into $D$-dimensional vectors. A 1D convolution layer is shown as (kernel size, /stride, \#in channels$\to$\#out channels, group number).}	   
	\label{Fig.2} 
\end{figure}

Based on previous studies, it is necessary to divide the time series into patches before embedding\cite{nie2022time,zhang2023crossformer}. And it is also common in 2D convolution-based models to split the original input into patches at the network’s beginning for downsampling \cite{liu2022convnet,he2015deep}. Therefore, we adopt a patch-style embedding strategy in our 1D convolution-based model to divide the input time series into patches and embed them. 

As shown in Figure \ref{Fig.2}, we have ${\mathbf{X}_{in}}\in {{\mathbb{R}}^{M\times L}}$ as the $M$ variables input time series of length $L$. In the patching process, the time series is divided into $N$ patches of patch size $P$. And the stride in the patching process is $S$, which also serves as the length of non overlapping region between two consecutive patches. Technically, after unsqueezing its shape to ${\mathbf{X}_{in}}\in {{\mathbb{R}}^{M\times 1\times L}}$, we can divide the time series into $N$ patches and then embed them to $D$-dimensional vectors by using a 1D convolution stem layer as follows:
\begin{equation}\label{eq4}
\mathbf{X} _{emb} = \mathrm{Conv1d} (\mathrm{Padding} (\mathbf{X} _{in} ))_{kernel\ size=P,stride=S,channels:1\to D} 
\end{equation}
Where ${\mathbf{X}_{emb}}\in {{\mathbb{R}}^{M\times D\times N}}$ is the input embedding.  $\mathrm{Padding}(\cdot )$ denotes the padding operation applied to the original time series $\mathbf{X} _{in}$ to keep the number of patches $N=L//S$. Specifically, we repeat $\mathbf{X} _{in}$'s last value $(P-S)$ times and then pad them back to the end of $\mathbf{X} _{in}$ before patching and embedding. For $\mathrm{Conv1d}(\cdot )$ here, we set its kernel size as $P$ and stride as $S$. And it maps 1 input channel into D output channels.
The overall embedding process in Section \ref{sec32} can also be briefly summarized by Equation \ref{eq1}.

\subsection{Cross-LKTCN Block}
\label{sec33}
\begin{figure}[htb]
	\centering 
	\includegraphics[width=0.95\textwidth]{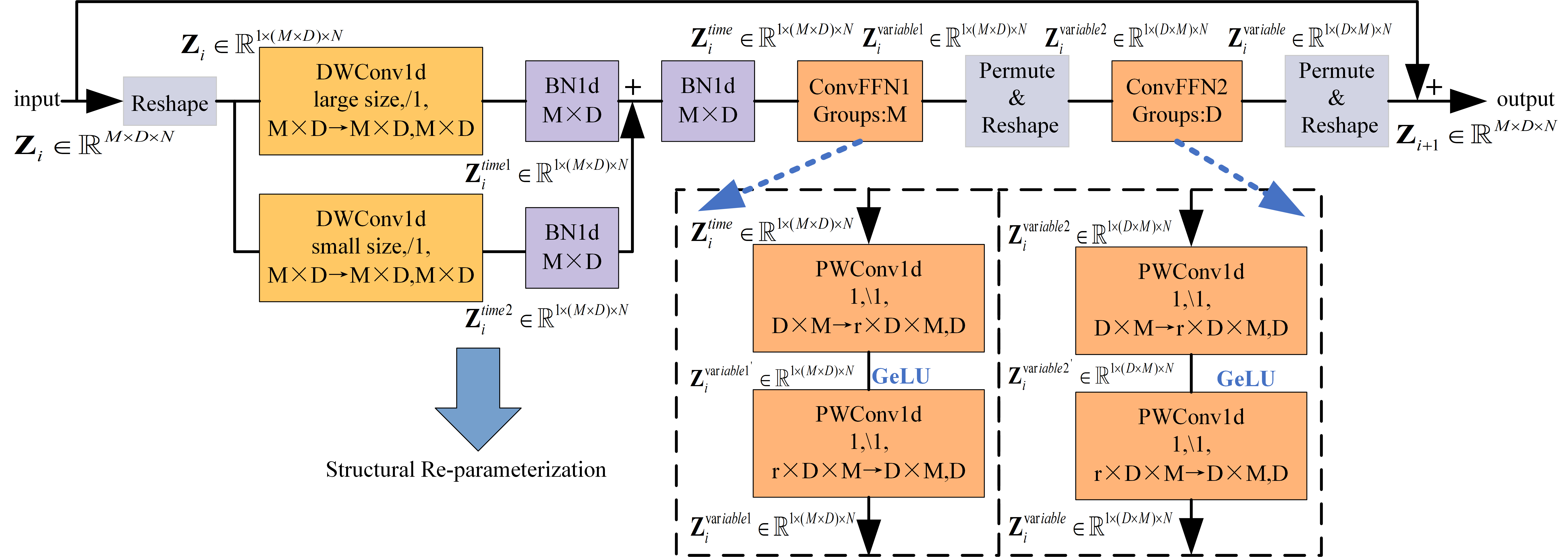}
	\caption{Structure of Cross-LKTCN block. It contains a depth-wise large kernel convolution to capture cross-time dependency and two successive point-wise group convolution FFNs to capture the cross-variable dependency. A 1D convolution layer is shown as (kernel size, /stride, \#in channels$\to$\#out channels, group number). DW and PW are short for depth-wise and point-wise.}	   
	\label{Fig.3} 
\end{figure}

Cross-time dependency and cross-variable dependency are critical for multivariate time series forecasting. To this end, we propose Cross-LKTCN block (Figure \ref{Fig.3} ) as a basic module to capture both cross-time and cross-variable dependency in time series. Therefore the backbone containing $K$ Cross-LKTCN blocks can learn informative representation from the input embedding $\mathbf{X}_{emb}$. Cross-LKTCN block is organized in a residual way \cite{he2015deep}. The forward process in the $i$-th Cross-LKTCN block is:
\begin{equation}\label{eq81}
\mathbf{Z} _{i+1} = \mathrm{Block} (\mathbf{Z} _{i}) + \mathbf{Z} _{i}
\end{equation}

Where ${\mathbf{Z}_{i}}\in {{\mathbb{R}}^{M\times D\times N}}$ , $i \in \{1,...,K\}$ is the $i$-th block’s input,
\begin{equation}\label{eq8}
\mathbf{Z} _{i} =\left\{
\begin{aligned}
&\mathbf{X} _{emb} &,i=1   \\ 
&\mathrm{Block} (\mathbf{Z} _{i-1}) + \mathbf{Z} _{i-1} &,i>1 
\end{aligned}
\right.
\end{equation}
$\mathrm{Block}(\cdot )$ denotes the Cross-LKTCN block. In details, a Cross-LKTCN block contains a depth-wise large kernel convolution to capture long-term temporal dependency and two successive point-wise group convolution FFNs to capture cross-variable dependency. Therefore, the detailed forward process in the $i$-th Cross-LKTCN block can be introduced in the following paragraphs.

\paragraph{Depth-wise Large Kernel Convolution.}
Firstly, we merge the first two axes of the $i$-th block’s input and unsqueeze its shape to  ${\mathbf{Z}_{i}}\in {{\mathbb{R}}^{1\times (M\times D)\times N}}$. Then a depth-wise large kernel convolution is used to capture cross-time dependency:

\begin{equation}\label{eq9}
\begin{split}
\mathbf{Z} _{i}^{time1}=&\ \mathrm{BN} (\mathrm{DW1Conv1d} (\mathbf{Z} _{i})) _{kernel\ size=large\ size,channels:(M\times D)\to (M\times D),groups=(M\times D)} \\
\mathbf{Z} _{i}^{time2}=&\ \mathrm{BN} (\mathrm{DW2Conv1d} (\mathbf{Z} _{i})) _{kernel\ size=small\ size,channels:(M\times D)\to (M\times D),groups=(M\times D)} \\
\mathbf{Z} _{i}^{time}=&\ \mathbf{Z} _{i}^{time1}+\mathbf{Z} _{i}^{time2}
\end{split}
\end{equation}
$\mathbf{Z}_{i}^{time}\in {{\mathbb{R}}^{1\times (M\times D)\times N}}$ is the representation after capturing the cross-time dependency. $\mathrm{BN}(\cdot )$ means 1D Batch Normalization\cite{ioffe2015batch}. $\mathrm{DW1Conv1d}(\cdot )$ and $\mathrm{DW2Conv1d}(\cdot )$ are two parallel depth-wise convolution layers to map $(M\times D)$ input channels into  $(M\times D)$ output channels with different kernel sizes.
 Here we set the kernel size in $\mathrm{DW1Conv1d}(\cdot )$ as $large\ size$ to enlarge the receptive field.

 And we set the kernel size in $\mathrm{DW2Conv1d}(\cdot )$ as $small\ size$. $\mathrm{DW2Conv1d}(\cdot )$ serves as an additional Structural Re-parameterization branch which helps to make up the optimization issue of large kernel convolutions according to \cite{ding2022scaling,liu2022more}.
 More details about Structural Re-parameterization are in \textcolor{blue}{supplementary materials}. The $large\ size$ and $small\ size$ are two hyperparameters to be defined. 

\paragraph{Successive Point-wise Group Convolution FFNs.} We adopt two successive point-wise group convolution FFNs to capture cross-variable dependency. The process is as follows:
\begin{equation}\label{eq10}
\begin{split}
\mathbf{Z} _{i}^{variable1}=&\mathrm{ConvFFN1} (\mathbf{Z} _{i}^{time}) _{groups=M} \\
\mathbf{Z} _{i}^{variable2}=&\mathrm{Permute\&Reshape} (\mathbf{Z} _{i}^{variable1})\\
\mathbf{Z} _{i}^{variable}=&\mathrm{ConvFFN2} (\mathbf{Z} _{i}^{variable2}) _{groups=D} \\
\mathbf{Z} _{i+1}=&\mathrm{Permute\&Reshape} (\mathbf{Z} _{i}^{variable}) \\
\end{split}
\end{equation}
The former $\mathrm{Permute\&Reshape}(\cdot )$ after $\mathrm{ConvFFN1}(\cdot )$ changes the shape from $\mathbf{Z} _{i}^{variable1}\in {{\mathbb{R}}^{1\times (M\times D)\times N}}$ to $\mathbf{Z} _{i}^{variable2}\in {{\mathbb{R}}^{1\times (D\times M)\times N}}$.
And $\mathbf{Z} _{i}^{variable}\in {{\mathbb{R}}^{1\times (D\times M)\times N}}$  is the representation after capturing the cross-variable dependency. 
After the latter $\mathrm{Permute\&Reshape}(\cdot )$, we have $\mathbf{Z} _{i+1}\in {{\mathbb{R}}^{M\times D\times N}} $ as the output of the $i$-th Cross-LKTCN block. 
We set the group number of $\mathrm{ConvFFN1}(\cdot )$ as $M$ to learn the new $D$ features of each variable per time step and set the group number of $\mathrm{ConvFFN2}(\cdot )$ as $D$ to capture the cross-variable dependency per feature in each time step. Weights in $\mathrm{ConvFFN1}(\cdot )$ and $\mathrm{ConvFFN2}(\cdot )$ are shared among all time steps. 

And the forward process in $\mathrm{ConvFFN1}(\cdot )$ is as follows:
\begin{equation}\label{eq11}
\begin{split}
{\mathbf{Z} _{i}^{variable1}} ^{'} =&\mathrm{Drop}(\mathrm{PW} _{1}^{1}\mathrm{Conv1d} (\mathbf{Z} _{i}^{time})_{kernelsize=1,channels:(M\times D)\to (r\times M\times D),groups=M} )\\
{\mathbf{Z} _{i}^{variable1}} ^{'} =&\mathrm{GELU}({\mathbf{Z} _{i}^{variable1}} ^{'})\\
{\mathbf{Z} _{i}^{variable1}}  =&\mathrm{Drop}(\mathrm{PW} _{2}^{1}\mathrm{Conv1d} ({\mathbf{Z} _{i}^{variable1}}^{'} )_{kernelsize=1,channels:(r\times M\times D)\to (M\times D ),groups=M}) 
\end{split}
\end{equation}

Similarly, the forward process in $\mathrm{ConvFFN2}(\cdot )$ is as follows:
\begin{equation}\label{eq12}
\begin{split}
{\mathbf{Z} _{i}^{variable2}} ^{'} =&\mathrm{Drop}(\mathrm{PW} _{1}^{2}\mathrm{Conv1d} (\mathbf{Z} _{i}^{variable2})_{kernelsize=1,channels:(D\times M)\to (r\times D\times M),groups=D} ) \\
{\mathbf{Z} _{i}^{variable2}} ^{'} = &\mathrm{GELU}({\mathbf{Z} _{i}^{variable2}} ^{'})\\
{\mathbf{Z} _{i}^{variable}}  =&\mathrm{Drop}(\mathrm{PW} _{2}^{2}\mathrm{Conv1d} ({\mathbf{Z} _{i}^{variable2}}^{'} )_{kernelsize=1,channels:(r\times D\times M)\to (D\times M) ,groups=D}) 
\end{split}
\end{equation}
$\mathrm{GELU}(\cdot )$ is the non-linear activation \cite{hendrycks2016gaussian}. $\mathrm{Drop}(\cdot )$ is dropout operation\cite{srivastava2014dropout}. $\mathrm{PW} _{1}^{1}\mathrm{Conv1d}(\cdot )$ and $\mathrm{PW} _{2}^{1}\mathrm{Conv1d}(\cdot )$ are the first and the second point-wise group convolutions in $\mathrm{ConvFFN1}(\cdot )$. The same goes for $\mathrm{ConvFFN2}(\cdot )$.  And $r$ is the FFN ratio, which means in the FFN process, the channel number is mapping from $(D\times M)$ to $(r\times D\times M)$ and then mapping back to $(D\times M)$ with group convolutions.

\begin{figure}[htb]
	\centering 
	\includegraphics[width=0.85\textwidth]{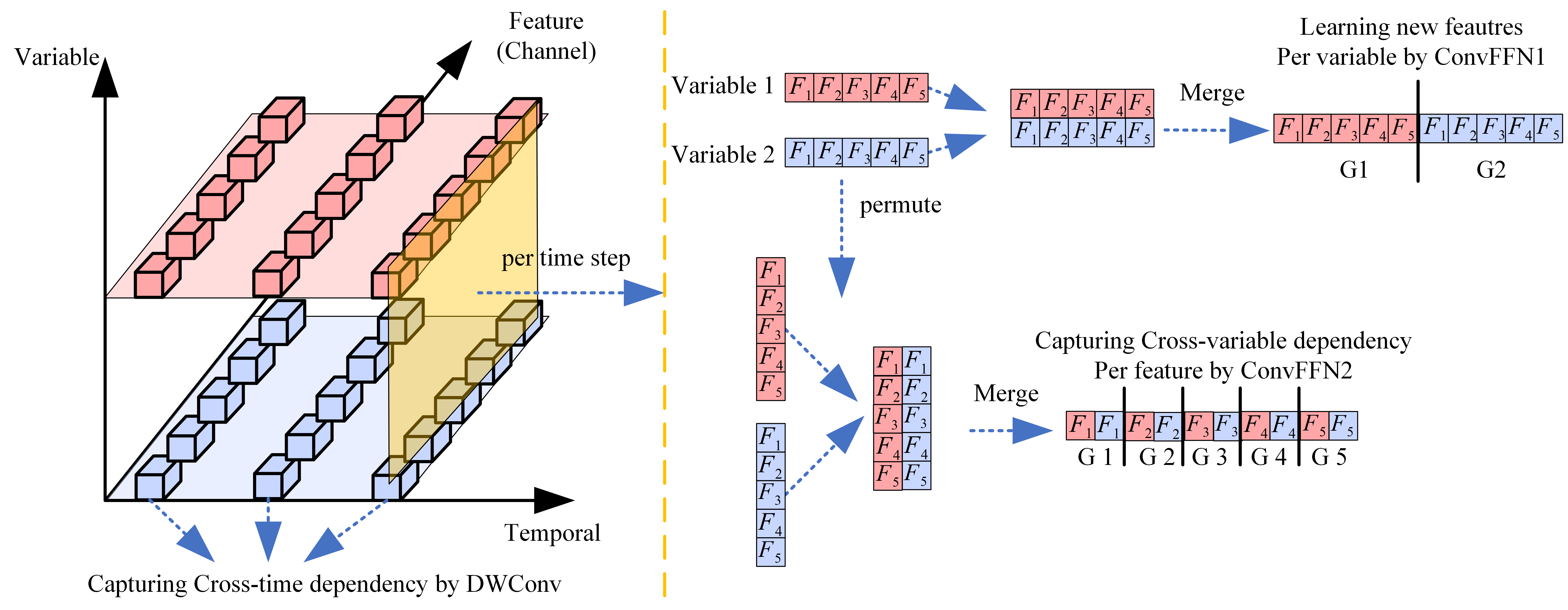}
	\caption{An example to describe the roles of each module in a Cross-LKTCN block. Here the input of Cross-LKTCN block contains three tokens with variable number as 2 and channel number as 5.}	   
	\label{Fig.4} 
\end{figure}

In conclusion, in a Cross-LKTCN block, the $\mathrm{DW1Conv1d}(\cdot )$ is used to capture cross-time dependency. The $\mathrm{ConvFFN1}(\cdot )$ is responsible for applying the linear transformation and non-linear activation to $D$ previous learned features per variable to further learn the new representation of each variable independently. And $\mathrm{ConvFFN2}(\cdot )$ is in charge of capturing the cross-variable dependency per feature. Details are shown in Figure \ref{Fig.4}.

\textbf{In the following process,} $\mathbf{Z}_{i+1}$ will be sent to the next Cross-LKTCN block as its input. And the output of the last Cross-LKTCN block $\mathbf{Z}_{K+1}=\mathrm{Block} (\mathbf{Z} _{K}) + \mathbf{Z} _{K}$ will be treated as the final representation $\mathbf{Z}$ generated by the backbone in Equation \ref{eq2}.

\section{Experiments}
\label{exps}

\paragraph{Datasets.} We evaluate the performance of our proposed Cross-LKTCN on 9 popular real-world datasets, including Weather, Traffic, Electricity, Exchange, ILI and 4 ETT datasets (ETTh1, ETTh2, ETTm1, ETTm2). More details about the datasets are described in \textcolor{blue}{supplementary materials}.

\paragraph{Baselines.}For multi-variate time series forecasting, we choose three state-of-the-art transformer-based models: PatchTST\cite{nie2022time}, FEDformer\cite{zhou2022fedformer} and Autoformer\cite{wu2021autoformer}, two recent cross-variable models: MTGNN\cite{wu2020connecting} and Crossformer\cite{zhang2023crossformer}, two latest convolution-based models: MICN\cite{wangmicn} and SCINet\cite{liu2022scinet}, and one state-of-the-art Linear model: DLinear\cite{zeng2022transformers} as our baselines. For some baselines that have different variants, we perform experiments on all the variants and collect the best results.

\paragraph{Implementation.} All models are following the same experimental setup with prediction length $T \in \{24, 36, 48, 60\}$ for ILI dataset and $T \in \{96, 192, 336, 720\}$ for other datasets as \cite{nie2022time}. We collect some baseline results from \cite{nie2022time}. Following the protocols in \cite{nie2022time}, for other baselines, we follow their official implementation but re-run them with various input length $L$ and choose the best results to avoid under-estimating the baselines.  All experiments are repeated three times. We calculate the MSE and MAE of multivariate time series forecasting as metrics. More details about the implementation are described in \textcolor{blue}{supplementary materials}.

\begin{table}
	\caption{Multivariate long-term forecasting results. We set the prediction lengths $T \in \{24, 36, 48, 60\}$ for ILI dataset and $T \in \{96, 192, 336, 720\}$ for other datasets. A lower MSE or
		MAE indicates a better performance. The best
		results are in \textbf{bold} and the second best are \underline{underlined}.}
	\label{table-result}
	\centering
	\begin{threeparttable}
		\resizebox{0.9\textwidth}{!}{

			\begin{tabular}{c|c|cc|cc|cc|cc|cc|cc|cc|cc|cc}
				
				\toprule
				
				
				\multicolumn{2}{c|}{Models}
				&\multicolumn{2}{c|}{Cross-LKTCN}
				&\multicolumn{2}{c|}{PatchTST}
				&\multicolumn{2}{c|}{DLinear}
				&\multicolumn{2}{c|}{Crossformer}
				&\multicolumn{2}{c|}{MTGNN}
				&\multicolumn{2}{c|}{MICN}
				&\multicolumn{2}{c|}{SCINet}
				&\multicolumn{2}{c|}{FEDfromer}
				&\multicolumn{2}{c}{Autoformer} \\

				\midrule
				
				\multicolumn{2}{c|}{Metric}
				&MSE &MAE
				&MSE &MAE
				&MSE &MAE
				&MSE &MAE
				&MSE &MAE
				&MSE &MAE
				&MSE &MAE
				&MSE &MAE
				&MSE &MAE\\

				\midrule
				
				\multirow{4}{*}{\rotatebox{90}{Electricity}}&96 &\textbf{0.129} &\underline{0.226} &\textbf{0.129} &\textbf{0.222}  &\underline{0.140} &0.237 &0.187 &0.283 &0.198 &0.294 &0.159 &0.267 &0.171 &0.256 &0.186 &0.302 &0.196 &0.313 \\
			
				&192 &\textbf{0.143} &\textbf{0.239} &\underline{0.147} &\underline{0.240} &0.153 &0.249 &0.258 &0.330 &0.266 &0.339 &0.168 &0.279 &0.177 &0.265 &0.197 &0.311 &0.211 &0.324 \\
			
				&336 &\textbf{0.161} &\textbf{0.259} &\underline{0.163} &\textbf{0.259} &\underline{0.169} &0.267 &0.323 &0.369 &0.328 &0.373 &0.196 &0.308 &0.197 &0.285 &0.213 &0.328 &0.214 &0.327 \\
			
				&720 &\textbf{0.191} &\textbf{0.286} &\underline{0.197} &\underline{0.290}  &0.203 &0.301 &0.404 &0.423 &0.422 &0.410 &0.203 &0.312 &0.234 &0.318  &0.233 &0.344 &0.236 &0.342\\

				\midrule
				
				\multirow{4}{*}{\rotatebox{90}{ETTh2}}&96 &\textbf{0.263} &\textbf{0.332} &\underline{0.274} &\underline{0.336}  &0.289 &0.353 &0.628 &0.563
				 &0.690 &0.614 &0.289 &0.357 &0.295 &0.361 &0.332 &0.374 &0.332 &0.368 \\
			
				&192 &\textbf{0.320} &\textbf{0.374} &\underline{0.339} &\underline{0.379} &0.383 &0.418 &0.703 &0.624 &0.745 &0.662 &0.409 &0.438 &0.349 &0.383 &0.407 &0.446 &0.426 &0.434 \\
			
				&336 &\textbf{0.313} &\textbf{0.376} &\underline{0.329} &\underline{0.380} &0.448 &0.465 &0.827 &0.675 &0.886 &0.721 &0.417 &0.452 &0.365 &0.409 &0.400 &0.447 &0.477 &0.479 \\
			
				&720 &\underline{0.392} &\underline{0.433} &\textbf{0.379} &\textbf{0.422} &0.605 &0.551 &1.181 &0.840 &1.299 &0.936 &0.426 &0.473 &0.475 &0.488  &0.412 &0.469 &0.453 &0.490\\

				\midrule

				\multirow{4}{*}{\rotatebox{90}{Weather}}&96 &\textbf{0.149} &\underline{0.204} &\textbf{0.149} &\textbf{0.198}  &0.176 &0.237 &\underline{0.153} &0.217 &0.161 &0.223&0.161 &0.226 &0.178 &0.233 &0.238 &0.314 &0.249 &0.329 \\
				
				&192 &\underline{0.196} &\underline{0.248} &\textbf{0.194} &\textbf{0.241}  &0.220 &0.282 &0.197 &0.269 &0.206 &0.278 &0.220 &0.283  &0.235 &0.277 &0.275 &0.329 &0.325 &0.370 \\
				
				&336 &\textbf{0.238} &\textbf{0.281} &\underline{0.245} &\underline{0.282}  &0.265 &0.319 &0.252 &0.311 &0.261 &0.322 &0.275 &0.328 &0.337 &0.345 &0.339 &0.377 &0.351 &0.391 \\
				
				&720 &\textbf{0.314} &\textbf{0.334} &\textbf{0.314} &\textbf{0.334}  &0.323 &\underline{0.362} &\underline{0.318} &0.363 &0.324 &0.366 &0.311 &0.356 &0.396 &0.413  &0.389 &0.409 &0.415 &0.426\\
				\midrule
				
				\multirow{4}{*}{\rotatebox{90}{ILI}}&24 &\underline{1.347} &\textbf{0.717} &\textbf{1.319} &\underline{0.754}  &2.215 &1.081 &3.040 &1.186 &4.268 &1.385 &2.684 &1.112 &2.150 &1.005 &2.624 &1.095 &2.906 &1.182 \\
				
				&36 &\textbf{1.250} &\textbf{0.778} &\underline{1.430} &\underline{0.834} &1.963 &0.963 &3.356 &1.230 &4.768 &1.494 &2.507 &1.013 &2.103 &0.983 &2.516 &1.021 &2.585 &1.038 \\
				
				&48 &\textbf{1.388} &\textbf{0.781} &\underline{1.553} &\underline{0.815} &2.130 &1.024 &3.441 &1.223 &5.333 &1.592 &2.423 &1.012 &2.432 &1.061 &2.505 &1.041 &3.024 &1.145 \\
				
				&60 &\underline{1.774} &\underline{0.868} &\textbf{1.470} &\textbf{0.788}  &2.368 &1.096 &3.608 &1.302 &5.083 &1.556 &2.653 &1.085 &2.325 &1.035  &2.742 &1.122 &2.761 &1.114\\
				\midrule
				
				\multirow{4}{*}{\rotatebox{90}{Traffic}}&96 &\underline{0.373} &\underline{0.263} &\textbf{0.360} &\textbf{0.249}  &0.410 &0.282 &0.512 &0.290 &0.527 &0.316 &0.508 &0.301 &0.613 &0.395 &0.576 &0.359 &0.597 &0.371 \\
				
				&192 &\underline{0.383} &\underline{0.257} &\textbf{0.379} &\textbf{0.256} &0.423 &0.287 &0.523 &0.297 &0.534 &0.320 &0.536 &0.315 &0.559 &0.363 &0.610 &0.380 &0.607 &0.382 \\
				
				&336 &\textbf{0.391} &\textbf{0.263} &\underline{0.392} &\underline{0.264} &0.436 &0.296 &0.530 &0.300 &0.540 &0.335 &0.525 &0.310 &0.555 &0.358 &0.608 &0.375 &0.623 &0.387 \\
				
				&720 &\underline{0.435} &\underline{0.288} &\textbf{0.432} &\textbf{0.286}  &0.466 &0.315 &0.573 &0.313 &0.557 &0.343 &0.571 &0.323 &0.620 &0.394  &0.621 &0.375 &0.639 &0.395\\
				\midrule
				
				\multirow{4}{*}{\rotatebox{90}{Exchange}}&96 &\underline{0.080} &\underline{0.196} &0.093 &0.214  &0.081 &0.203 &0.186 &0.346 &0.203 &0.381 &0.102 &0.235 &\textbf{0.061} &\textbf{0.188} &0.139 &0.276 &0.197 &0.323 \\
				
				&192 &0.166 &\underline{0.288} &0.192 &0.312 &\underline{0.157} &0.293 &0.467 &0.522 &0.459 &0.512 &0.172 &0.316 &\textbf{0.106} &\textbf{0.244} &0.256 &0.369 &0.300 &0.369 \\
				
				&336 &0.307 &\underline{0.398} &0.350 &0.432 &\underline{0.305} &0.414 &0.783 &0.721 &0.707 &0.697 &0.272 &0.407 &\textbf{0.181} &\textbf{0.323} &0.426 &0.464 &0.509 &0.524 \\
				
				&720 &0.656 &\underline{0.582} &0.911 &0.716 &\underline{0.643} &0.601 &1.367 &0.943 &1.323 &0.912 &0.714 &0.658 &\textbf{0.525} &\textbf{0.571}  &1.090 &0.800 &1.447 &0.941\\
				\bottomrule
				
		\end{tabular}}
		\begin{tablenotes}
			\item[1] More results about other ETT benchmarks are provided in \textcolor{blue}{supplementary materials}.  
		\end{tablenotes}
	\end{threeparttable}
\end{table}

\subsection{Main Results}
Table \ref{table-result} shows the multivariate long-term forecasting results. Overall, our model outperforms all baseline methods. We discuss the results from the following aspects. (1) \textbf{Compared with convolution-based models} like MICN\cite{wangmicn} and SCINet\cite{liu2022scinet}, Cross-LKTCN achieves an overall 27.4\% reduction on MSE and 15.3\% reduction on MAE, becoming the best performing convolution-based model. These experimental results indicate that our structure can better unleash the potential of convolution in time series forecasting. (2) \textbf{Compared with  cross-variable baselines} like MTGNN\cite{wu2020connecting} and Crossformer\cite{zhang2023crossformer}, Cross-LKTCN achieves an overall 52.3\% reduction on MSE and 33.5\% reduction on MAE, indicating our proposed method is more efficient in capturing cross-variable dependency. It’s worth noting that Cross-LKTCN also outperforms the previous state-of-the-art variable-independent methods in general, indicating the importance of cross-variable dependency capturing. (3) \textbf{Compared with state-of-the-art models}, Cross-LKTCN can still achieve comparable or even better performance. Cross-LKTCN can outperform DLinear\cite{zeng2022transformers} in general. Compared with PatchTST\cite{nie2022time}, Cross-LKTCN can surpass it in many datasets. For example, in Exchange dataset that is of unclear periodicity, while PatchTST suffers from performance degradation for it cannot capture useful cross-time information, Cross-LKTCN  can utilize cross-variable dependency as additiaonal information and still achieves top-2 perfomance, which is much better that PatchTST. And SCINet has advantages in Exchange dataset for its hierarchical structure can utilize multi-scale information of input series, which makes it more possible to match the actual periodicity. And in other settings, the prediction accuracy of Cross-LKTCN and PatchTST are very close. \textbf{The experimental results prove that a cross-variable model and a pure convolution-based model can also achieve state-of-the-art performance and have great potential in time series forecasting.} (4) See \textcolor{blue}{supplementary materials} for showcases. 

\subsection{Ablation Study}

\paragraph{Successive Point-wise Group Convolution FFNs.}
We propose two successive point-wise group convolution FFNs to capture cross-variable dependency. To validate its effectiveness, we consider following 4 cases: (a) \emph{D Groups + M Groups.} means our two successive point-wise group convolution FFNs whose group numbers equal to M and D respectively. (b) \emph{M Groups.} means we only keep the first point-wise group convolution FFN whose group number equals to M. (c) \emph{D Groups.} means we only keep the second point-wise group convolution FFN whose group number equals to D. (d) \emph{no Group.} means we replace the two successive point-wise group convolution FFNs with one non-group convolution FFN. Results are shown on Table \ref{table-ffn}. Case (a) achieves the best results. As comparison, case (b) can only apply the linear transformation and non-linear activation to $D$ features per variable to learn each variable's deep representation independently. But it omits to capture the cross-variable dependency. In contrast, case (c) can capture the cross-variable dependency per feature. But to each variable, it doesn't apply any linear interaction across features to learn the new deep representation. And case (d) is not a decoupling method, it cannot meet the needs of capturing cross-variable dependency and learning the new deep representation of each variable respectively.

\begin{table}
	\caption{Ablation of successive point-wise group convolution FFNs. We compare our method with three different variants on three datasets. The best results are highlighted in \textbf{blod}.}
	\label{table-ffn}
	\centering
	\resizebox{0.9\textwidth}{!}{

		\begin{tabular}{cc|c|cccc|cccc|cccc}
			\toprule
			\multicolumn{3}{c|}{Datasets}& \multicolumn{4}{c|}{ILI} & \multicolumn{4}{c|}{ETTh1} &\multicolumn{4}{c}{Electricity} \\
			
			\midrule
			\multicolumn{3}{c|}{Prediction length} &24&36&48&60 &96&192&336&720 &96&192&336&720\\
			
			\midrule
			\multicolumn{2}{c|}{\multirow{2}{*}{\emph{M Groups + D Groups.}}} & MSE & \textbf{1.347} &\textbf{1.250} &\textbf{1.388} &\textbf{1.774} &\textbf{0.368} &\textbf{0.405} &\textbf{0.391} &\textbf{0.450} &\textbf{0.130} &\textbf{0.143} &\textbf{0.161} &\textbf{0.191} \\
			& &MAE &\textbf{0.717} &\textbf{0.778} &\textbf{0.781} &\textbf{0.868} &\textbf{0.394} &\textbf{0.413} &\textbf{0.412} &\textbf{0.461} &\textbf{0.226} &\textbf{0.239} &\textbf{0.261} &\textbf{0.286}\\
			
			\midrule
			\multicolumn{2}{c|}{\multirow{2}{*}{\emph{M Groups.}}} & MSE &1.858 &1.269 &1.658 &1.883 &0.375 &0.426 &0.403 &0.467 &0.135 &0.151 &0.168 &0.199 \\
			& &MAE &0.895 &0.789 &0.876 &0.902 &0.403 &0.429 &0.423 &0.483 &0.233 &0.249 &0.272 &0.300\\
			
			\midrule
			\multicolumn{2}{c|}{\multirow{2}{*}{\emph{D Groups.}}} & MSE &2.525 &1.318 &1.662 &1.991 &0.377 &0.412 &0.409 &0.467 &0.136 &0.153 &0.170 &0.197 \\
			& &MAE &0.935 &0.730 &0.907 &0.957 &0.406 &0.429 &0.438 &0.479 &0.235 &0.250 &0.276 &0.298\\
			
			\midrule
			\multicolumn{2}{c|}{\multirow{2}{*}{\emph{no Group.}}} & MSE &2.231 &1.482 &1.551 &1.991 &0.381 &0.423 &0.418 &0.479 &0.138 &0.161 &0.173 &0.203 \\
			& &MAE &0.906 &0.796 &0.873 &0.926 &0.412 &0.427 &0.432 &0.489 &0.238 &0.254 &0.278 &0.304\\
			
			\bottomrule
			
	\end{tabular}}
\end{table}

\subsection{Model Analysis}

\paragraph{Impact of Kernel Size.}
\label{ap_ks}
According to\cite{ding2022scaling}, a large kernel size is the key to obtain a large receptive field in 2D convolution. To verify whether this finding still works on 1D convolution and to figure out the impact of kernel size, we perform experiments with 3 different kernel sizes ranging from small to large on 3 datasets. Results on Table \ref{table-ks} show that increasing the kernel size leads to performance improvement. The experiment results indicate that directly enlarging the kernel size in 1D convolution layer and training it with Structural Re-parameterization technique\cite{ding2022scaling} can effectively improve the receptive field and help convolution layer to better capture cross-time dependency.

\begin{table}
	\caption{Impact of kernel size. We compare three different kernel sizes ranging from small to large. A lower MSE or MAE indicates a better performance. The best results are highlighted in \textbf{blod}.}
	\label{table-ks}
	\centering
	\resizebox{0.9\textwidth}{!}{

		\begin{tabular}{cc|c|cccc|cccc|cccc}
			\toprule
			\multicolumn{3}{c|}{Datasets}& \multicolumn{4}{c|}{ILI} & \multicolumn{4}{c|}{ETTh1} &\multicolumn{4}{c}{Electricity} \\
			
			\midrule
			\multicolumn{3}{c|}{Prediction length} &24&36&48&60 &96&192&336&720 &96&192&336&720\\
			
			\midrule
			\multicolumn{2}{c|}{\multirow{2}{*}{kernel size = 3}} & MSE &1.906 &1.546 &1.754 &1.893 &0.381 &0.416 &0.403 &0.460 &0.143 &0.155 &0.175 &0.203 \\
			& &MAE &0.862 &0.841 &0.891 &0.900 &0.405 &0.423 &0.419 &0.470 &0.237 &0.249 &0.270 &0.299\\
			
			\midrule
			\multicolumn{2}{c|}{\multirow{2}{*}{kernel size = 31}} & MSE &1.687 &1.486 &1.376 &1.855 &\textbf{0.367} &0.405 &\textbf{0.389} &\textbf{0.449} &0.133 &0.147 &\textbf{0.159} &0.192 \\
			& &MAE &0.848 &0.855 &0.797 &0.929 &\textbf{0.393} &0.413 &\textbf{0.410} &\textbf{0.460} &0.228 &0.243 &\textbf{0.257} &0.288\\
			
			\midrule
			\multicolumn{2}{c|}{\multirow{2}{*}{kernel size = 51}} & MSE &\textbf{1.347} &\textbf{1.250} &\textbf{1.388} &\textbf{1.774} &0.368 &\textbf{0.403} &0.391 &0.449 &\textbf{0.130} &\textbf{0.143} &0.161 &\textbf{0.189}\\
			& &MAE &\textbf{0.717} &\textbf{0.778} &\textbf{0.781} &\textbf{0.868}  &\textbf{0.393} &\textbf{0.412} &0.411 &\textbf{0.460} &\textbf{0.226} &\textbf{0.240} &0.260 &\textbf{0.286}\\
			
			\bottomrule
			
	\end{tabular}}
\end{table}

\begin{figure}[htb]
	\centering 
	\includegraphics[width=0.95\textwidth]{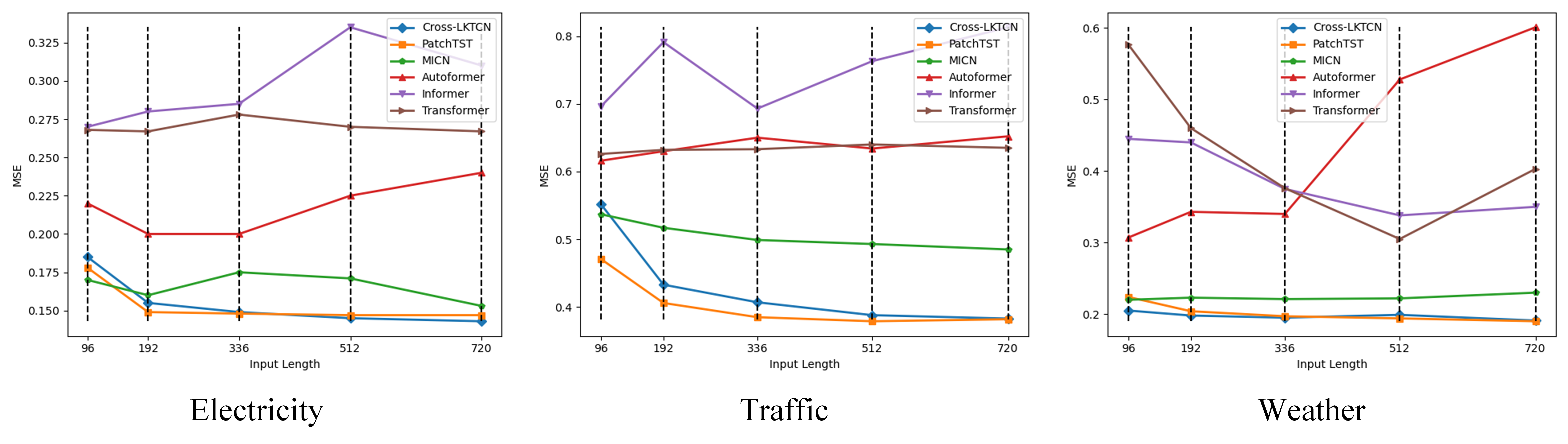}
	\caption{The MSE results with different input lengths and same prediction length(192 time steps).}	   
	\label{fig-inputlength} 
\end{figure}

\paragraph{Impact of Input Length.}
\label{ap_inputlength}
Since a longer input length indicates more historical information an algorithm can utilize in time series forecasting, a model with strong ability to capture long-term temporal dependency should perform better when input length increases\cite{zeng2022transformers,wangmicn,nie2022time}. To validate our model, we conduct experiments with different input lengths and the same prediction length. As shown in Figure \ref{fig-inputlength},in general, our model gains performance improvement with increasing input length, indicating our model can effectively extract useful information from longer history and capture long-term dependency. However, some Transformer-based models\cite{wu2021autoformer,zhou2021informer,vaswani2017attention} suffer from performance degradation with increasing input length owing to the repeated short-term patterns according to\cite{zhou2021informer}.

\subsection{Efficiency Analysis}

We list each layer's complexity to capture cross-time and cross-variable dependency in two latest cross-variable/convolution-based models on Table \ref{table-complexity1} for comparison.
 For Crossformer\cite{zhang2023crossformer}, we only analyse the complexity in the attention module in an Encoder block. And for MICN\cite{wangmicn}, since it is a variable-mixing method, we don't take variable number $M$ into consideration when analysing its complexity and don't consider its complexity to capture cross-variable dependency.
Cross-LKTCN and Crossformer's complexity to capture cross-variable dependency are quadratic with respect to variable number $M$. But in terms of capturing cross-time dependency, while Crossformer suffers from quadratic complexity due to the attention mechanism, Cross-LKTCN can reduce the computational cost to linear complexity.

Cross-LKTCN and MICN's complexity to capture cross-time dependency are linear with respect to input length $L$ since they are pure convolution structure. But with the help of patch-style embedding, the coefficient $\frac{1}{S}$ term can significantly reduce Cross-LKTCN's practical complexity.  More running time and memory usage analysis are in \textcolor{blue}{supplementary materials}.
\begin{table}
	\caption{ Complexity analysis of different forecasting models.}
	\label{table-complexity1}
	\centering
	\resizebox{0.4\textwidth}{!}{

		\begin{tabular}{c|cc}
			\toprule
			Model& Cross-time & Cross-variable  \\
			
			\midrule
			Cross-LKTCN &$O(\frac{L}{S}MD)$ &$O(\frac{L}{S}MD^{2}+\frac{L}{S}DM^{2})$\\
			
			\midrule
			Crossformer &$O(\tfrac{L^{2}}{S^{2}}MD)$ & $O(M^{2}\frac{L}{S}D)$\\

			\midrule
			MICN &$O(LD^{2})$ & - \\

			\bottomrule
			
	\end{tabular}}
\end{table}

\section{Conclusion and Future Work}
This paper proposes a pure convolution-based model, namely cross-LKTCN, as an efficient way to capture cross-time and cross-variable dependency for time-series forecasting. Specifically, the patch-style embedding strategy is applied to time series to enhance locality and aggregate more semantic information. Then  in each Cross-LKTCN block, a depth-wise  large kernel convolution is used to capture cross-time dependency, following which two successive point-wise group convolution FFNs are used to capture cross-variable dependency. Extensive experimental results on nine real-world datasets demonstrate the effectiveness of Cross-LKTCN against state-of-the-arts. Our study also demonstrates the importance of cross-variable dependency and the effectiveness of convolution in time series forecasting. Both two aspects are worth further study in the future.

\newpage
\bibliographystyle{plain}
\bibliography{submissionref}

%


\end{document}